# Control of synaptic plasticity via the fusion of reinforcement learning and unsupervised learning in neural networks[1]

*Mohammad Modiri*[2]

**Abstract**

The brain can learn to execute a wide variety of tasks quickly and efficiently. Nevertheless, most of the mechanisms that enable us to learn are unclear or incredibly complicated. Recently, considerable efforts have been made in neuroscience and artificial intelligence to understand and model the structure and mechanisms behind the amazing learning capability of the brain. However, in the current understanding of cognitive neuroscience, it is widely accepted that synaptic plasticity plays an essential role in our amazing learning capability. This mechanism is also known as the Credit Assignment Problem (CAP) and is a fundamental challenge in neuroscience and Artificial Intelligence (AI). The observations of neuroscientists clearly confirm the role of two important mechanisms including the error feedback system and unsupervised learning in synaptic plasticity. With this inspiration, a new learning rule is proposed via the fusion of reinforcement learning (RL) and unsupervised learning (UL). In the proposed computational model, the nonlinear optimal control theory is used to resemble the error feedback loop systems and project the output error to state of neurons, and an unsupervised learning rule based on state of neurons or activity of neurons are utilized to simulate synaptic plasticity dynamics to ensure that the output error is minimized.

---

[1] This article is part of a new learning rule that shared to get feedback from researchers.
[2] Email address: Mohammad.Modiri@Gmail.com

# 1. Proposed framework

In this section, we propose a novel brain-inspired learning rule. The main idea is to combine a RL-based method and unsupervised learning to control synaptic plasticity. Fig. 1 shows the main ideas underlying this computational model that can be applied in different tasks such as classification, prediction, and robotics.

In the proposed computational model, learning law reformulated as an optimal control problem. For this purpose, all tasks even classification assumed as a dynamic system that illustrated in Fig. 1 (C). Since the NNs can be considered a special class of high-dimensional nonlinear dynamic systems, where both state and time are continuous. Therefore, continuous-time formulation of the HJB equation and ADP can be applied to derive the learning rule (optimal control law). The derived learning rule forces the proposed framework's output to mimic the reference trajectory that implicitly minimized the cross-entropy. The following subsections elaborate on the network architecture, and the learning rule in further detail.

## 1.1 Network architecture and dynamics

The proposed computational model consists of a feedforward and feedback pathway. The feedforward pathway can be any form of neural network, however a three parts module composed of an input layer, a neural network (NN), and an output layer is recommended. The feedback pathway also different types of Actor-Critic or Critic only can be employed, although an Actor-Critic is recommended.

Fig. 1 shows the block diagram of the proposed computational model architecture. The functionality of the proposed architecture is based on the following steps:

a. The encoding layer projects input patterns $x(t)$ into the NN.
b. In the pertaining process, a supervised learning algorithm like Gradient descent, Ridge, etc., can be employed to adjust the decoder parameters associated with each training sample. Once the decoder is trained, it is considered constant during the NN training process. (Optional)
c. In the NN training process. An RL-based method and unsupervised learning are proposed in subsection 1.2 is applied to make the NN learn temporal relations between the input patterns $x(t)$ and desired output $y(t)$ by adjusting the state (membrane potential) of neurons and consequently parameters in the NN using a novel learning rule. In the proposed error feedback mechanism, two RC-based actor and critic are utilized to approximate true state of neurons and NN parameters.

d. Model recall on new data.

The feedforward pathway resembles the structure of traditional neural networks, nevertheless the approach of the feedback pathway particularly proposed in this study differs considerably from classical artificial intelligence methods [1].

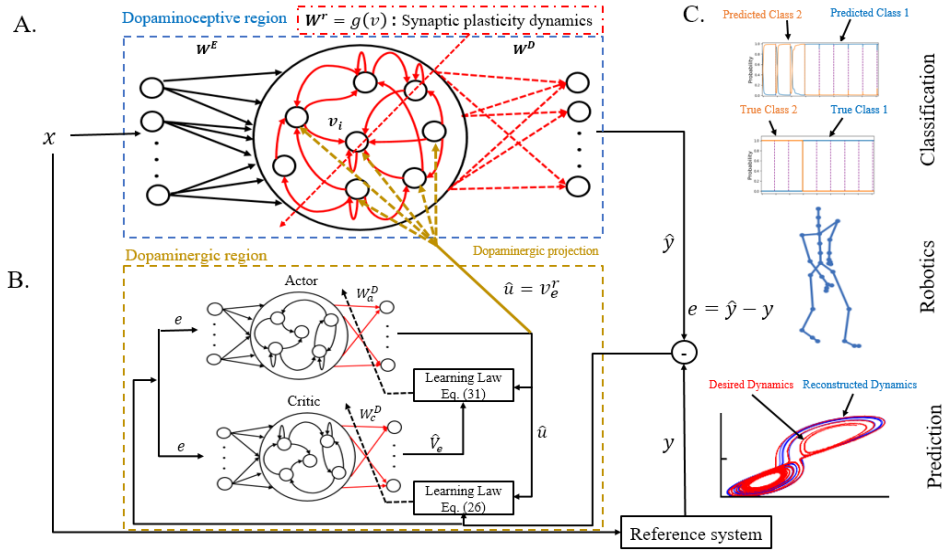

**Fig. 1.** A schematic of the proposed architecture (plastic weights in red). (A) The feedforward pathway consists of an encoding layer with fix and random parameters, a NN with trainable parameter and a decoder layer, which is trained in the pre-training process and stay constant during the NN training process. (B) In the feedback pathway, the nonlinear optimal control is applied for estimating state (membrane potential) of neurons and consequently NN parameters. It consists of actor and critic neural networks. (C) The subplots shows the reference trajectories and learned trajectories in difference tasks.

**Remark 1** For the sake of brevity, time dependence is suppressed while denoting variables of dynamical systems. For instance, the notations $x(t), y(t), u(t), v(t)$ and $e(t)$ are rewritten as $x, y, u, v$ and $e$. Moreover, all math symbols used in this study are listed in Table 1.

**Table 1** The math symbols used in this paper and corresponding meanings

| Symbol | Meaning |
| --- | --- |
| $n^r$ | Number of neurons |
| $N^r$ | Number of connections |
| $x$ | Input pattern |
| $y$ | Desired output |
| $\hat{y}$ | Estimated output |
| $e$ | Output error |
| $W^E$ | Encoder connections |
| $W^D$ | Decoder connections |
| $W^r$ | Network connections |
| $v_d$ | Desirable state (membrane potential) of neuron |
| $v_e$ | Difference between desirable and estimated state (membrane potential) of neuron |
| $q^E$ | filtered spike activity of input in spiking NN and is equal to input in non-spiking NN |
| $q^r$ | filtered spike activity of reservoir in spiking NN and is equal to state in non-spiking neurons |
| $S$ | spike train in spiking NN |
| $\psi(\cdot)$ | Leaky term |
| $\phi(\cdot)$ | Activation function |
| $\ell(\cdot,\cdot)$ | Utility function |
| $f(\cdot)$ | Drift dynamic function |
| $g(\cdot)$ | Input dynamic |
| $\mathcal{H}(\cdot,\cdot,\cdot)$ | Hamiltonian function |
| $u^*$ | Desirable state (membrane potential) of neuron |
| $u$ | Admissible policy |
| $u_2$ | Estimated state (membrane potential) of neuron (input control) |
| $V^*$ | Optimal cost function |
| $V$ | Cost function |
| $V_v$ | Derivative of the cost function |
| $\hat{W}_1$ | Critic weights |
| $\hat{W}_2$ | Actor weights |
| $\alpha_1$ | Learning rate of the actor |
| $\alpha_2$ | Learning rate of the critic |

As illustrated in Fig. 1, The core element of the proposed computational model is a neural network with the adjacency matrix $W^r \in \mathbb{R}^{n^r \times n^r}$. Here, the NN consists of $n^r$ neurons, for which the membrane potential dynamics are described as:

$$\dot{v}_i = \psi(v_i) + \mathrm{I}_i \tag{1}$$

$$I_i = W_i^E \emptyset(v^E) + W_i^r \emptyset(v) \qquad (2)$$

, where $v = [v_1. v_2. \cdots . v_{n^r}] \in \mathbb{R}^{n^r}$ is the state (membrane potential) of neuron of the NN neurons, $\emptyset(\cdot)$ is a nonlinear dendrite, and, $\psi(v): \mathbb{R}^{n^r} \to \mathbb{R}^{n^r}$ is a leak-term where we used $-\alpha_l v$, $v^E$, $v^r$ in non-spiking neurons are directly equal to input $x$ and NN state $v$, and in spiking neurons $v_i$ shows the filtered spike activity of neuron $i$. $S_i(t)$ is the spike train of the neuron $i$ and modelled as a sum of Dirac delta-functions:

$$q_i = (S_i * \kappa)(t) = \int_{-\infty}^{t} S_i(s)\kappa(t-s)ds \qquad (3)$$

$$\kappa(t) = \exp(-t/\tau)/\tau \qquad (4)$$

In additional, a linear decoder is supposed as:

$$\hat{y} = h(v) = W^D v \qquad (5)$$

### 1.2 Synaptic plasticity mechanism

The equations mentioned in section 1.1 shows that the NN synaptic plasticity mechanism can be modeled as a dynamical system. Thus, principles of optimal control theory can be utilized to derive a learning rule (optimal control law). For this purpose, we reformulate the AI tasks as a control problem. Therefore, the output error were considered as follows:

$$e = \hat{y} - y \in \mathbb{R}^c \qquad (6)$$

, where $y \in \mathbb{R}^c$ denotes the desired output at time $t$, $\hat{y} \in \mathbb{R}^c$ is the corresponding predicted output, and $c$ is a number of output variables. By derivation of Eq. (6) with respect to $t$, the error dynamics can be written as:

$$\dot{e} = \dot{\hat{y}} - \dot{y} \in \mathbb{R}^c \qquad (7)$$

Thus, the error dynamics only depend on the NN neural dynamics is equal to:

$$\dot{e} = W^D \dot{v} - W_d^D \dot{v}_d$$
$$= W^D[\psi(v(t)) + W^E v^E(t) + W^r(t)v(t)] \quad (8)$$
$$- W_d^D[\psi_d(v(t)) + W_d^E v^E(t) + W_d^r(t)v_d(t)]$$

**Assumption 1** Assume the following simplification conditions:

$$W_d^E = W^E$$
$$W_d^D = W^D \quad (9)$$

According to Assumption 1, the error dynamics only depend on the NN.

$$\dot{e} = W^D[\psi_e(v(t)) + W^r(t)v_f(t)] = W^D \dot{v}_e \quad (10)$$

We assumed:

$$\psi_e(v(t)) = f(v(t))$$
$$W^r(t) = g(v(t)) \quad (11)$$
$$u^r(t) = v(t) - v_d(t) = v_f(t)$$

Therefore, neural space Eq. (8) can be rewritten as the following affine state space equation from:

$$\dot{e} = W^D \dot{v}_e = W^D[f(v(t)) + g(v(t))u^r(t)] \quad (12)$$

Where the control input is considered as state of neurons changes $\dot{v}(t)$ and $w^r$ are function of control input of the NN, and $u^r$, $W^r$ which the number are selected according to the problem:

$$u^r = [u^f; u] \in \mathbb{R}^{n^r} \quad (13)$$

$$W^r = [W^f; W] \in \mathbb{R}^{n^r \times n^r} \quad (14)$$

, in $u \in \mathbb{R}^{n^r-n^f}$ and $W \in \mathbb{R}^{(n^r-n^f) \times (n^r-n^f)}$ is fixed or even zero NN parameters, $u \in \mathbb{R}^n$ and $W \in \mathbb{R}^{n \times n}$ is the NN control input and synaptic plasticity respectively and $[x; y]$ means combination of $x$ and $y$.

In the artificial intelligence (AI) tasks, the goal of learning is to adapt the NN's parameters $W^r$ such that the error is minimized and the NN's parameters and control input remains bounded. Thus, the following cost function was considered:

$$\min_{u(\cdot) \in U} \Im(e(\cdot). u(\cdot)) = \int_t^{t_f} \ell(e(\tau). u(\tau)) d\tau \tag{15}$$

$$\ell(e. u) = e^T Q e + u^T R u \approx \ell(v_e. u) = v_e^T Q v_e + u^T R u \tag{16}$$

, wherein $\ell(\cdot. \cdot)$ is the utility function, $t_f$ is the final time or last element of input patterns, $Q. R$ symmetric positive definite matrix for ensuring that the error in cost function is sufficiently affective. Hence, the optimal value function can be written:

$$V(e) \approx V(v_e) = \inf_{u(\cdot) \in U} \Im(u. v_e) \tag{17}$$

Ultimately, it is ideal to achieve an optimal input control $u^*$ or $v_e^*$ and weight update law $g^*(v)$ that stabilizes the system Eq. (10) and minimizes the cost function Eq. (17). This kind of input control $u$ is called admissible control [2].

Now, the learning rule has been reformulated given the error dynamic in Eq. (10) and the cost function Eq. (17). To solve this dynamic optimization problem, the HJB equation is utilized, so the Hamiltonian of the cost function Eq. (17) associated with control input $u$ is defined as:

$$\mathcal{H}(v_e. u. V_e) = \ell(v_e. u) + V_e^T (\dot{v}_e) \tag{18}$$

, where $V_e = \partial V / \partial e$ is the partial derivative of the cost function, for admissible control policy $\mu$ we have:

$$\mathcal{H}(v_e. \mu. V_e) = 0 \tag{19}$$

The present study assumed that the solution to Eq. (19) is smooth giving the optimal cost function:

$$V^*(e) \approx V^*(v_e) = \min_u \left( \int_t^{t_f} \ell(u(\tau).v_e(\tau)) d\tau \right) \qquad (20)$$

, which satisfies the HJB equation

$$\min_u \mathcal{H}(v_e.u.V_e^*) = 0 \qquad (21)$$

Assuming that the minimum on the left hand side Eq. (21) exists then by applying stationary condition $\partial \mathcal{H}(v_e.u.V_e)/\partial u = 0$, the learning rule (optimal control) can be obtained as:

$$u^*(e) \approx u^*(v_e) = -\frac{1}{2} R^{-1} g^T(v) V_e^* \qquad (22)$$

The optimal value function can be obtained as:

$$V^*(v_e) = \min_u \left( \int_t^{t_f} v_e^T Q v_e + u^{*T} R u^* d\tau \right) \qquad (23)$$

Inserting this optimal learning rule Eq. (22) into nonlinear Lyapunov equation Eq.(18) gives the formulation of the HJB equation Eq. (21) in terms of $V_e^*$

$$0 = v_e^T Q v_e + V_e^{*T}(v_e) f(v) - \frac{1}{4} V_e^{*T}(v_e) g(v) R^{-1} g^T(v) V_e^*(v_e) \qquad (24)$$

Finding the error feedback mechanism to stabilize the NN requires solving the HJB equation for the value function and then substituting the solution to obtain the desired state of neurons update and consequently learning rule. Although HJB gives the necessary and sufficient condition for optimality of a learning rule (control law) with respect to a loss function; Unfortunately, according to the nonlinear characteristics of the NN, solving the HJB equation in explicit form is difficult or even impossible to derive for systems of interest in practice. However, a few solution such as Feynman-Kac lemma, Al'brecht, Garrard, Viscosity, Games based, Adaptive Dynamic Programming (ADP) methods have been proposed to solve HJB approximately. Meanwhile, ADP offers a data driven solution that is more homogeneous with proposed framework. Thus, the proposed computational focuses on the ADP method to approximate its solution. To address this mater, with the inspiration of the dopaminergic region structure and conformity with the Weierstrass high-order approximation theorem [3], we utilized a new actor-critic based on the seminal Reservoir Computing (s-RC) that proposed in [4]. Briefly, the objective of tuning the actor weights is to minimize the approximate value, and the objective of tuning the critic weights is to minimize the Bellman equation error.

Therefore, a s-RC is used as critic to approximate the derivatives of the value function:

$$V_e(e) \approx V_{v_e}(v_e) = W_c^D z_c + \varepsilon_c \in \mathbb{R}^{c \times 1}$$
$$z_c \in \mathbb{R}^{n_c \times 1}$$
$$W_c^D \in \mathbb{R}^{c \times n_c}$$
(25)

Where $W_c^D$ is the decoder weight, and $z_c$ is a representation of input $v_e$. By applying the gradient descent algorithm, critic weight update rule will be:

$$\dot{\widehat{W}}_c^D = -\alpha_c \frac{\partial E_c}{\partial \widehat{W}_c^D} = -\alpha_c (f(v) + g(v)u) z_c^T e_c$$
$$= -\alpha_c \sigma_c (\sigma_c^T \widehat{W}_c^D + v_e^T Q v_e + u^T R u)$$
(26)

Let $\widehat{W}_c^D$ is an estimation of unknown matrix $W_c^D$, $\alpha_c > 0$ is critic learning rate, and $\sigma_c = (f(v) + g(v)u) z_c^T$.

Similarly, one more s-RC is utilized as the actor to approximate the state (membrane potential) of neurons:

$$u(v_e) = W_a^D z_a + \varepsilon_a \in \mathbb{R}^{N \times 1}$$
$$z_a \in \mathbb{R}^{n_a \times 1}$$
$$W_a^D \in \mathbb{R}^{N \times n_a}$$
(27)

Where $z_a$ be the new representation of input $v_e$, $\widehat{W}_a^D$ an estimation of unknown matrix $W_a^D$. By using the gradient descent algorithm, a weight update expression for the actor can be written as follows:

$$\dot{\widehat{W}}_a^D = -\alpha_a \frac{\partial E_a}{\partial \widehat{W}_a^D} = -\alpha_a (\widehat{W}_a^D z_a + \frac{1}{2} R^{-1} g^T(v)(\widehat{W}_c^D z_c)) z_a^T$$
(28)

Where $\alpha_a > 0$ is actor learning rate.

Consequently, in the course of the NN learning process, the feedback pathway forces the output of the computational model to follow the reference trajectory, an effect that is widely used in control theory. After an adequate learning time, the feedforward pathway in the absent of feedback pathway can fallow desirable patterns with acceptable accuracy.

The stability and convergence of the proposed framework and error can be investigated similar to [4]. However, there is an important issue that should be noted, where unlike the [4], the $g(\cdot)$ is not clearly specified. Although, varied candidate solutions that can be unsupervised learning such as the bounded spike timing dependent plasticity (STDP) can be suggested. Nevertheless, a comprehensive investigation requires for providing an ideal solution for $g(\cdot)$, which is ignored in the current version of article.